\documentclass[10pt,twocolumn,letterpaper]{article}

\usepackage[pagenumbers]{cvpr} 

\usepackage[dvipsnames]{xcolor}

\definecolor{cvprblue}{rgb}{0.21,0.49,0.74}
\usepackage[pagebackref,breaklinks,colorlinks,citecolor=cvprblue]{hyperref}
\usepackage{color,array}
\usepackage{xcolor}
\usepackage{colortbl}
\usepackage{graphicx}
\usepackage{amsmath}
\usepackage{amssymb}
\usepackage{booktabs}
\usepackage{multirow}
\usepackage{bm}
\usepackage[pagebackref,breaklinks,colorlinks]{hyperref}

\usepackage[capitalize]{cleveref}
\usepackage{bm}
\newtheorem{prop}{Proposition}

\crefname{section}{Sec.}{Secs.}
\Crefname{section}{Section}{Sections}
\Crefname{table}{Table}{Tables}
\crefname{table}{Tab.}{Tabs.}

\def\paperID{8998} 
\def\confName{CVPR}
\def\confYear{2024}

\title{FINER: Flexible spectral-bias tuning in Implicit NEural Representation \\by Variable-periodic Activation Functions}

\author{
Zhen Liu$^{1,\dagger}$, Hao Zhu$^{1,\dagger}$, Qi Zhang$^2$, 
Jingde Fu$^3$, Weibing Deng$^{3}$, Zhan Ma$^{1}$, Yanwen Guo$^{4}$, 
Xun Cao$^{1}$\\
$^1$ {School of Electronic Science and Engineering}, 
$^3$ {Department of Mathematics}, \\
$^4$ {Department of Computer Science and Technology},  
Nanjing University, Nanjing 210023, China\\
$^2$ AI Lab, Tencent Company, Shenzhen 518054, China\\
$^\dagger$ Equal contribution.
Corresponding author: {\tt caoxun@nju.edu.cn}
}

\begin{document}
\maketitle

\begin{abstract}
Implicit Neural Representation (INR), which utilizes a neural network to map coordinate inputs to corresponding attributes, is causing a revolution in the field of signal processing. However, current INR techniques suffer from a restricted capability to tune their supported frequency set, resulting in imperfect performance when representing complex signals with multiple frequencies. We have identified that this frequency-related problem can be greatly alleviated by introducing variable-periodic activation functions, for which we propose FINER. By initializing the bias of the neural network within different ranges, sub-functions with various frequencies in the variable-periodic function are selected for activation. Consequently, the supported frequency set of FINER can be flexibly tuned, leading to improved performance in signal representation. We demonstrate the capabilities of FINER in the contexts of 2D image fitting, 3D signed distance field representation, and 5D neural radiance fields optimization, and we show that it outperforms existing INRs.
\end{abstract}

\begin{figure} 
    \begin{center}
        \includegraphics[width=\linewidth]{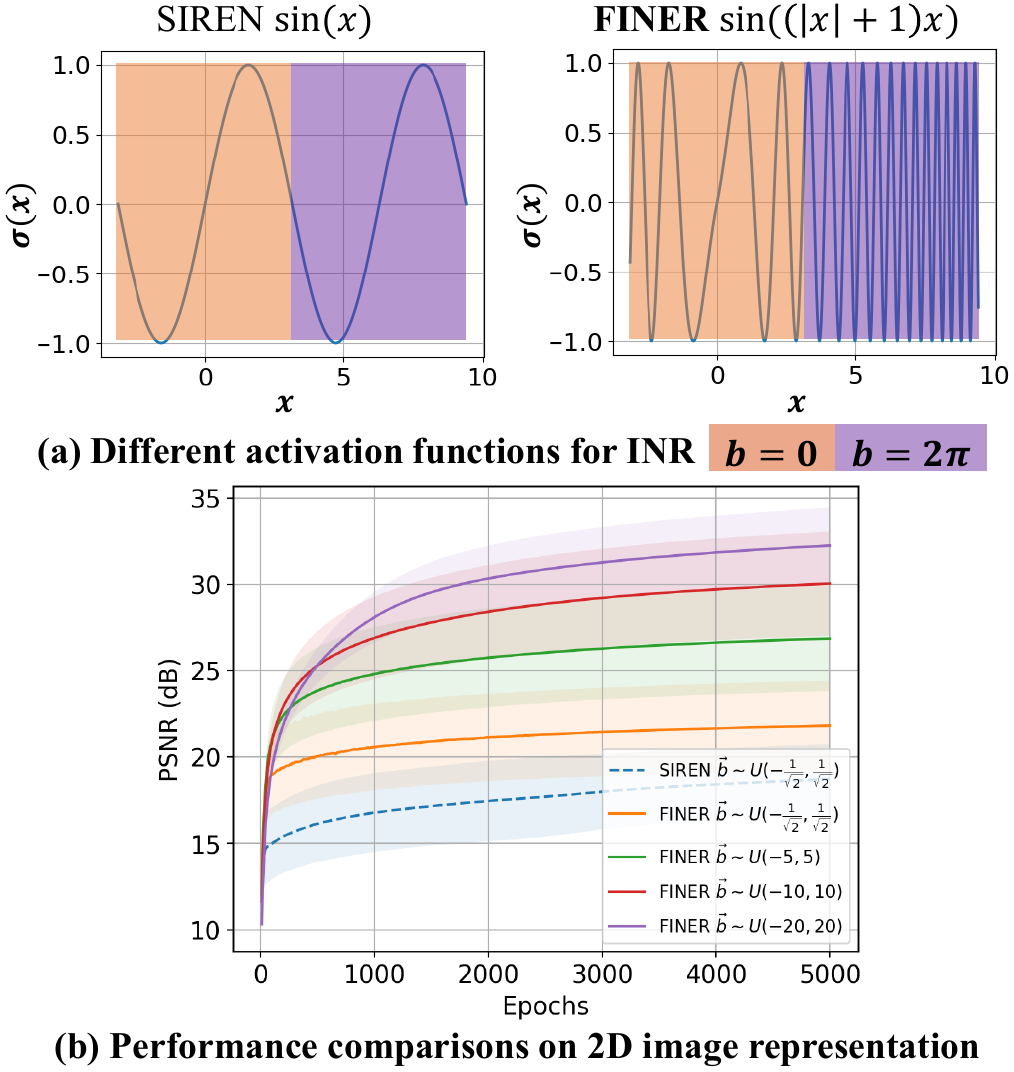}
        \caption{\textbf{Flexible spectral-bias tuning in implicit neural representation (FINER)}. We observe that the supported frequency set in classical INRs is limited due to the underutilization of activation functions' definition domain, \textit{i.e.}, they mainly employ the central region near the origin point. To overcome this limitation, we propose a novel variable-periodic activation function for INR. This innovation allows for tuning the supported frequency set by adjusting the initialization range of the bias vector in the neural network. (a) visualizes the selected narrow activation functions in classical periodic activation $\sin(x)$ alongside our proposed variable-periodic one $\sin((|x|+1)x)$ with different bias settings. (b) plots the training curves of SIREN and FINER, demonstrating the impact of different initializations applied to the bias vector $\vec{b}$ (see Sec.~\ref{sec:img_diff_bias} for more details).}
        \label{fig:firstimg}
    \end{center}
    \vspace{-1cm}
\end{figure} 

\section{Introduction}
The way a signal is represented is the foundation of all the following problems and determines the paradigm for solving them. Traditional representations, such as the image matrices, point cloud or volumes~\cite{tewari2022advances}, focus on recording the elements individually and have offered significant contributions in history. However, this representation is increasingly inadequate for addressing the numerous inverse problems arising in modern times, such as neural rendering~\cite{tewari2022advances}, inverse imaging~\cite{xie2022neural} and simulations~\cite{karniadakis2021physics}. On the contrary, the implicit neural representation (INR)~\cite{sitzmann2020implicit}, which characterizes a signal by preserving the mapping function from the coordinates to corresponding attributes using neural networks, is gaining increasing attentions thanks to the advantages of querying attributes continuously and incorporating differentiable physical process seamlessly. As a result, INR has found widespread application in solving domain-specific inverse problems~\cite{karniadakis2021physics,hao2022physics}, particularly in cases where large-scale paired datasets are unavailable, and only measurements and forward physical process are provided.

However, existing INR techniques suffer the well-known spectral-bias~\cite{rahaman2019spectral,yuce2022structured}, \textit{i.e.}, the low-frequency components of the signal are more easily to be learned. To address this bias, the positional encoding~\cite{mildenhall2020nerf,tancik2020fourier} which aims at embedding multiple orthogonal Fourier or Wavelet bases~\cite{fathony2020multiplicative} into the subsequent network is proposed. However, a significant challenge arises from the fact that the frequency distribution of a signal to be inversely solved is often unknown. This can potentially lead to a mismatch between the pre-defined bases' frequency set and the characteristics of the signal itself, resulting in an imperfect representation~\cite{xie2023diner}. Apart from the pre-defined frequency set, there is a growing focus within the research community on automatic frequency tuning, achieved through the use of periodic~\cite{sitzmann2020implicit} or non-periodic activation~\cite{ramasinghe2022beyond,saragadam2023wire} functions. Nevertheless, the supported frequency set is still limited and the representational accuracy could be further improved. 

Such a problem is closely related to the underutilization of the used activation functions. While activation functions have an infinite domain, practical applications typically make use of only the regions centered around the origin. 
By `interspersing' narrow activation functions with different frequencies along the full definition domain and then selecting the ideal one by controlling the range of input values, the supported frequency set could be significantly expanded, resulting in improved expressive power of current INRs. Following this idea, we propose the FINER, where variable-periodic functions are applied to activate the neurons within the INR.

Different from previous explorations~\cite{sitzmann2020implicit,ramasinghe2022beyond,saragadam2023wire} which focus on optimizing the weight matrix for manipulating frequency candidates with better matching degree, FINER opens a novel way to achieve frequency tuning by modulating the bias vector, or in other words, the phase of the variable-periodic activation functions. 
We demonstrate that both the shift-invariance and eigenvalues distribution of FINER's neural tangent kernel (NTK) can be enhanced (see Figs~\ref{fig:firstimg}, \ref{fig:NTK_vis}) by increasing the standard deviation during bias initialization, thus the spectral bias could be flexibly tuned and the expressive power are significantly improved.
To verify the performance, extensive experiments are conducted on 2D image fitting, 3D signed distance field representation and 5D neural radiance field optimization. Specifically, the main contributions of the work include,
\begin{enumerate}
    \item We introduce a novel implicit neural representation with flexible spectral-bias tuning for representing and optimizing signals.
    \item We propose a novel initialization scheme and prove its effectiveness and efficiency from both the perspectives of geometry and neural tangent kernel.
    \item We substantiate that FINER surpasses prior INRs activated with other functions for 2D image fitting, 3D signed distance field representation and 5D neural radiance field optimization.
\end{enumerate}

\section{Related Work}
\subsection{Implicit Neural Representation}
INRs~\cite{sitzmann2020implicit, tancik2020fourier} serve as the foundational building blocks for neural scene representations. These representations are designed to learn continuous functions using a multi-layer perception (MLP) that maps coordinates to visual signals, such as images~\cite{dupont2021coin, lindell2021bacon, xie2023diner}, videos~\cite{kasten2021layered, chen2021nerv}, and 3D scenes~\cite{mildenhall2020nerf, wang2021neus}. Neural Radiance Fields or NeRF \cite{mildenhall2020nerf}, a notable breakthrough in this domain, learns a 5D INR to reconstruct the radiance fields (density and view-dependent color) of a scene. With the widespread application of NeRF and its variants \cite{muller2022instant, barron2021mip} on realistic view synthesis, INRs have rapidly expanded into various fields of vision and signal processing, such as cross-model media representation/compression \cite{gao2022objectfolder, strumpler2022implicit}, neural camera representations \cite{huang2022hdr, huang2023inverting}, microscopy imaging \cite{zhu2022dnf, liu2022recovery} and partial differential equations solver~\cite{raissi2019physics, karniadakis2021physics}.
Despite the interest and success of these implicit representations, current approaches often suffer from the well-known spectral-bias problem. As a result, the INRs may struggle to capture high-fidelity details of complex signals, leading to suboptimal performance in fitting functions and ineffectiveness in various applications.

\subsection{Spectral-bias Problem}
The issue of spectral-bias in deep learning \cite{rahaman2019spectral} refers to the innate propensity of these models to selectively learn specific patterns or features from input data. In the case of INR-based methods, this problem manifests as a preference for learning low-frequency components of a signal more readily than high-frequency components. To address the spectral-bias problem, several innovative strategies have been proposed for INR-based methods. In particular, spatial encoding is applied to the input data, such as frequency or polynomial decomposition \cite{tancik2020fourier, singh2023polynomial, raghavan2023neural}, high-pass filtering \cite{wu2023neural, fathony2020multiplicative}, to emphasize high-frequency components before feeding into the model. \cite{tancik2020fourier} use the Fourier features mapped from spatial coordinates as the input of MLPs to improve the performance of INRs in adequately expressing high-frequency information of signals. Additionally, various architectural modifications have been integrated into INRs, including multi-scale or pyramid representations \cite{lindell2021bacon, saragadam2022miner, zhu2023pyramid}, which can aid in capturing both low-frequency and high-frequency components of a scene. \cite{lindell2021bacon} implement a multi-scale network architecture with a band-limited Fourier spectrum to minimize artifacts during the downsampling or upsampling process. However, the frequency distribution of a signal requiring inverse solving is often unknown, making it difficult to design an appropriate representation or model for the signal without prior knowledge of its frequency content.


In addition to the positional encoding with pre-defined frequency bases, there has been a growing interest in the research community for automatic frequency tuning through the use of nonlinearity activation functions \cite{sitzmann2020implicit, chng2022gaussian, ramasinghe2022beyond, saragadam2023wire}.
\cite{sitzmann2020implicit} propose the Sinusoidal Representation Network (SIREN), a method designed to represent complex signals and functions using periodic activation functions, especially sine functions. SIREN has demonstrated its effectiveness in representing intricate details and high-frequency components when compared to traditional activation functions like ReLU or sigmoid.
However, it is important to note that SIREN may necessitate careful initialization and hyperparameter tuning to achieve optimal performance. The periodic nature of sine functions can lead to oscillations and slower convergence, posing challenges in the training process. Furthermore, the size of the embedding space is limited, as the frequency distribution bias remains constant for different inputs, making it difficult to represent the diversity of frequency distributions. As a result, it is crucial to develop an adaptive periodic function for activating nonlinearity and complex frequency distribution.

\section{FINER: Flexible spectral-bias tuning in Implicit NEural Representation}
This section first provides an overview of SIREN, where the capacity-convergence gap is summarized. Then the FINER is proposed in detail. 

\subsection{SIREN and the Capacity-Convergence Gap}

\noindent \textbf{Pipeline of SIREN.} Given a signal sequence $\{\vec{x}_i,\vec{y}_i\}_{i=1}^{N}$, where $\vec{x}_i$ and $\vec{y}_i$ respectively represent the coordinate and the corresponding attributes of the $i$-th element, $N$ is the number of elements in the signal. SIREN focuses on pursuing a neural network $f(\vec{x};\theta)$ to characterize them as accurate as possible and could be formulated as follows:
\begin{equation}
    \begin{split}
        \displaystyle
        {\vec{z}}^{\:0} &= \vec{x} \\
        \vec{z}^{\:l} &= \sin(\omega_0(W^{l} \vec{z}^{\:l-1} + \vec{b}^{\:l})),\ l=1,2,...,L-1, \\
        f(\vec{x} ; \theta) &= W^{L} \vec{z}^{\:L-1} + \vec{b}^{\:L}
    \end{split}
    \label{eqn:periodic_activation}
    \vspace{-0.6cm}
\end{equation}
where $\displaystyle \vec{z}^{\:l}$ denotes the output of layer $l$, $\theta=\{W^{l}, \vec{b}^{\:l}\ |\ l=1,2,...,L-1\}$ refers to the network parameters to be optimized, $L$ is the number of layers, $\omega_0$ is an experienced parameter for controlling the frequency. 

\vspace{0.1in}
\noindent \textbf{Capacity-Convergence Gap.} According to \cite{yuce2022structured}, the capacity of the SIREN is limited by the choice of $\omega_0$, \textit{i.e.}, the functions that could be represented by $f(\vec{x};\theta)$ should be able to be decomposed by a linear combination of certain harmonics of $\omega_0$, \textit{i.e.},
\begin{equation}
    f(\vec{x};\theta) \in \left\{ \sum_{\omega'\in \mathcal{F}_{\omega_0}}c_{\omega'} \sin (\omega'\vec{x}+\phi_{\omega'}) \ | \ c_{\omega'},\phi_{\omega'} \in Q \right\}
\end{equation}
where $Q$ is the set of rational number, $\mathcal{F}_{\omega_0}$ is the supported frequency set defined by the $\omega_0$. Because $\omega_0$ only plays the role of scaling neurons, it could be removed by changing the initialization range of network parameters $\{W^{l},\vec{b}^{l} | \ l=1,...,L-1\}$. As a result, it is unnecessary to introduce the parameter $\omega_0$ in theory. 

On the other hand, since the $\sin$ function is non-convex, the network parameters should be initialized carefully to guarantee the inputs of $\sin$ mainly fall into the range of $[-\pi,\pi]$~\cite{sitzmann2020implicit}, otherwise SIREN will not be converged well and does not have a high accuracy for representing signals. For example, $W$ is often initialized from $\mathcal{U}(-\sqrt{6/n},\sqrt{6/n})$, $n$ is the number of inputs for each neuron~\cite{sitzmann2020implicit}. As a result, network parameters $\{W^{l},\vec{b}^{\:l} | \ l=1,...,L-1\}$ could not be scaled unboundedly considering the convergence. In summary, there is a capacity-convergence gap in SIREN that,
\begin{prop}
\label{prop:capacity_gap}
The function set that SIREN could be represented can be enlarged by increasing the initialization range of network parameters, which violates the rule of guaranteeing the convergence, resulting a performance gap between theory and practice.
\end{prop}

\begin{figure} 
    \begin{center}
        \includegraphics[width=\linewidth]{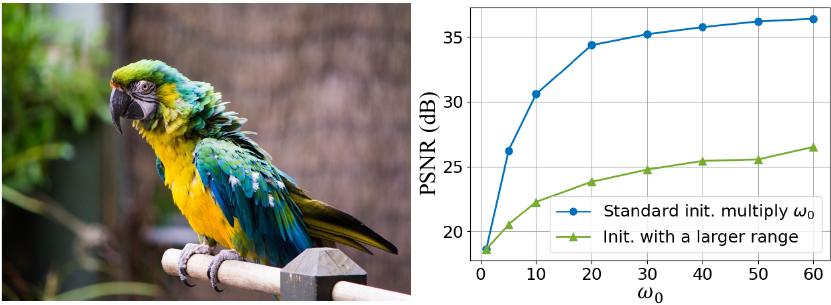}
        \caption{Comparisons of SIREN for fitting a $2K$ image. There is a performance gap between the results of ``Standard initialization multiply $\omega_0$'' and ``Initialization with a larger range''.}
        \label{fig:sec_3.1}
    \end{center}
    \vspace{-0.7cm}
\end{figure}

\noindent \textbf{A simple example: fitting a high-resolution image.} Fig.~\ref{fig:sec_3.1} compares the performance of SIREN with different initialization strategies. The blue line are the results follows the standard form of SIREN, \textit{i.e.}, $W\sim\mathcal{U}(-\sqrt{6/n},\sqrt{6/n})$ and an additional angular frequency $\omega_0$ is applied to multiply the $W\vec{x}+\vec{b}$. 

On the contrary, the green line refers to the performance by removing the angular frequency $\omega_0$ in Eqn.~\ref{eqn:periodic_activation} and initializing $W \sim \mathcal{U}(-\omega_0\sqrt{6/n},\omega_0\sqrt{6/n})$. Although they are equivalent in the mathematical form, there is a performance gap in practice.

\subsection{Variable-periodic Activation Functions}
To overcome the capacity-convergence gap, SIREN introduces the parameter $\omega_0$ to enlarge the supported frequency set. However, $\omega_0$ is often set manually which poses challenges for novices and requires additional tuning process for inverse problems where the frequency distribution of the signal to be solved is unknown. To address this problem, we propose the FINER which introduces a variable-periodic activation function $\sin((|x|+1)x)$, and the Eqn.~\ref{eqn:periodic_activation} could be rewritten as,
\begin{equation}
    \begin{split}
    \vec{z}^{\:l} &= \sin (\omega_0 \alpha^{l}(W^{l}\vec{z}^{\:l-1}+\vec{b}^{\:l})),\ l=1,2,...,L-1\\
    \alpha^{l}&=|W^{l}\vec{z}^{\:l-1}+\vec{b}^{\:l}|+1
    \end{split}.
    \label{eqn:var_activation}
\end{equation}


\noindent \textbf{More utilization of activation function closes the capacity-convergence gap.}
We notice that the periodic activation function $\sin(x)$ has an infinite definition domain while only the regions centered around the origin point are used. This underutilization of definition domain is caused by the periodic behavior of $\sin$. 
As shown in Fig.~\ref{fig:firstimg}(a), because $\sin(x)$ has a period $2\pi$, the behavior after activation will not be changed when shifting the used regions to other areas, such as orange and purple boxes in Fig.~\ref{fig:firstimg}(a) where $\vec{b}=0$ and $2\pi$, respectively. 
As a result, it is unnecessary to increase the utilization rate of the definition domain in SIREN where only the central region $[-2\pi, 2\pi]$ is used~\cite{sitzmann2020implicit}. Different from the periodic function $\sin(x)$, FINER adopts a variable-periodic function $\sin((|x|+1)x)$. Because the scale parameter $|x|+1$ increases with the increase of input variable, the frequency of the used activation function will be changed when the used regions are shifted, \textit{i.e.}, different initialization of network parameters $\vec{b}$. As shown in Fig.~\ref{fig:firstimg}(a), $\sin((|x|+1)x)$ for $\vec{b}=2\pi$ has higher frequencies than itself when $\vec{b}=0$. Thus the capacity-convergence gap could be closed by the proposed variable-periodic activation function.

\section{Flexible Spectral-bias Initialization}
In this section, we will first give the initialization scheme of FINER, then the behaviors of FINER's supported frequency set and training dynamics under different initialization are analysed from geometrical and neural tangent kernel perspectives, respectively.
\label{sec:ini_scheme}
\subsection{Initialization scheme} 
\begin{figure} 
    \begin{center}
        \includegraphics[width=\linewidth]{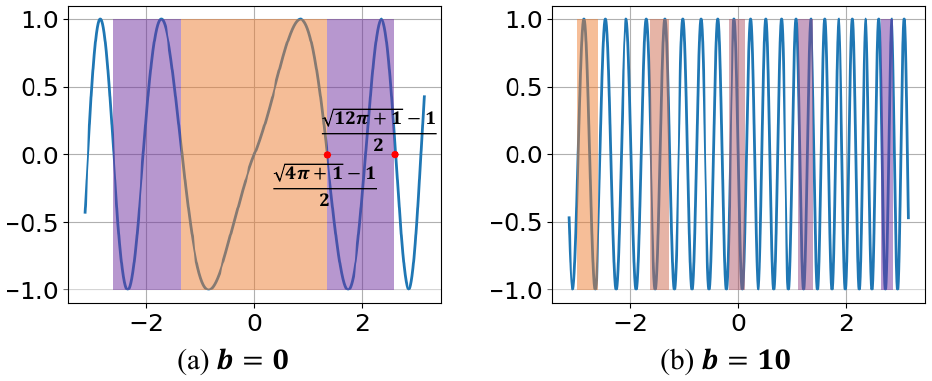}
        \vspace{-0.6cm}
        \caption{Comparisons of used activation function $\sin((|x|+1)x)$ under different bias $\vec{b}$. More sub-functions with high-frequency are included when $b$ is set with a larger value.}
        \label{fig:sec_3.3}
    \end{center}
    \vspace{-0.8cm}
\end{figure}

As demonstrated in the above section, the supported frequency set of FINER could be manipulated by changing the bias $\vec{b}$. However, due to the non-convex nature of the variable-periodic activation function, there are many local minimums in $\sin((|x|+1)x)$ and gradient-based optimizations (\textit{e.g.}, SGD or Adam) could not guarantee moving $\vec{b}$ to global optimum if $\vec{b}$ is not initialized well. Traditional initialization strategies, which apply uniform or Gaussian sampling in a narrow region centered at $0$, makes the supported frequency set of FINER be limited by the frequency of the first cycle in variable-periodic function (\textit{i.e.}, the Fig.~\ref{fig:sec_3.3}(a)), resulting in a waste of other cycles (Fig.~\ref{fig:sec_3.3}(b)) in variable-periodic function which have difference frequencies.

To get rid of the limited supported frequency set using traditional initialization methods, we derive a novel initialization scheme for $\vec{b}$ for tuning the supported frequency set flexibly, meanwhile the initialization for $W$ follows \cite{sitzmann2020implicit}. We propose to initialize $\vec{b}$ following the uniform distribution $\mathcal{U}(-k,k)$ with a larger range $k$ than the default one in traditional methods, 
\begin{equation}
    \vec{b} \sim \mathcal{U}(-k,k), \ k > 0.
\vspace{-0.2cm}
\end{equation}

\subsection{Geometrical Perspective} 
Supposing the supported frequency set of SIREN and FINER are $\mathcal{F}_{\omega_0}$ and $\mathcal{F}_{\omega_0,k}$, respectively. To analyse their relationship, let us start from the simplest case. 

\noindent \textbf{$k$ is close to the origin point.} Because the initialization for $W$ follows \cite{sitzmann2020implicit}, the term $W\vec{x}$ has similar distribution with the one in SIREN, that $|W\vec{x}|\leq \pi$. As a result, the term $|(|W\vec{x}+\vec{b}|+1)(W\vec{x}+\vec{b})|$ drops into the area of $[-\pi^2-\pi,\pi^2+\pi]$. As shown in Fig.~\ref{fig:sec_3.3}(a), the activation function $\sin((|W\vec{x}+\vec{b}|+1)(W\vec{x}+\vec{b}))$ mainly spans two narrow sub-functions with different frequencies. For the points dropped into the first sub-function (\textit{i.e.}, $|W\vec{x}|\leq\frac{\sqrt{4\pi+1}-1}{2}$, the orange areas in Fig.~\ref{fig:sec_3.3}(a)), the supported frequency set $\mathcal{F}_{\omega_0,k}^{1}$ is slightly larger than $\mathcal{F}_{\omega_0}$, and could be approximated as
\begin{equation}
    \mathcal{F}_{\omega_0,k}^{1} \approx \left\{ \frac{2\pi}{\sqrt{4\pi+1}-1} \omega \ | \ \omega\in \mathcal{F}_{\omega_0} \right\}, 
\end{equation}
where the term $\frac{2\pi}{\sqrt{4\pi+1}-1}$ measures the frequency changes from the standard $\sin$ function to the first function in $\sin((|x|+1)x)$. Considering the fact that the range $[-\pi, \pi]$ of $W\vec{x}$ is continuous, \textit{i.e.}, every value could be produced, the $\mathcal{F}_{\omega_0}$ is also a continuous set, as a result,
\begin{equation}
\label{eqn:siren_in_finer}
    \mathcal{F}_{\omega_0} \subset \mathcal{F}_{\omega_0,k}^{1}.
\end{equation}

For the points dropped into the second sub-function (\textit{i.e.}, $\frac{\sqrt{4\pi+1}-1}{2} \leq| W\vec{x}|\leq \frac{\sqrt{12\pi+1}-1}{2}$, the purple areas in Fig.~\ref{fig:sec_3.3}(a)), the supported frequency set $\mathcal{F}_{\omega_0,k}^{2}$ differs from $\mathcal{F}_{\omega_0,k}^{1}$ since the base frequency of the used activation changes. Compared with the $\mathcal{F}_{\omega_0,k}^{1}$ in the first sub-function, the $\mathcal{F}_{\omega_0,k}^{2}$ in the second one could be approximated as,
\begin{equation}
    \mathcal{F}_{\omega_0,k}^{2} \approx \left\{\frac{\sqrt{4\pi+1}-1}{\sqrt{12\pi+1}-\sqrt{4\pi+1}}\omega\ | \ \omega\in\mathcal{F}_{\omega_0,k}^{1}\right\},
\end{equation}
where $\frac{\sqrt{4\pi+1}-1}{\sqrt{12\pi+1}-\sqrt{4\pi+1}}$ characterizes the scale of frequency changes from the first sub-function to the second one. As a result, the supported frequency set $\mathcal{F}_{\omega_0,k}$ for $k$ is close to the origin is,
\begin{equation}
    \mathcal{F}_{\omega_0,k} = \mathcal{F}_{\omega_0,k}^{1} \cup \mathcal{F}_{\omega_0,k}^{2}.
\label{eqn:freq_set_small_b}
\end{equation}

\noindent \textbf{$k$ is far away from the origin point.} For example, $\vec{b}$ is initialized as $10$. 
Because the frequencies of each sub-functions are further increased for $\vec{b}=10$ (in Fig.~\ref{fig:sec_3.3}(b), the frequency is increased from the orange box to purple box), the supported frequency set $\mathcal{F}_{\omega_0,k}$ is increased a lot compared with the one in Eqn.~\ref{eqn:freq_set_small_b}. In summary, 
\begin{prop}
    The supported frequency set $\mathcal{F}_{\omega_0,k}$ of FINER increases with the increase of the initialization range of $\vec{b}$, and the supported frequency set $\mathcal{F}_{\omega_0}$ of SIREN is a subset of $\mathcal{F}_{\omega_0,k}$.
\end{prop}


\subsection{Neural Tangent Kernel Perspective}
Neural tangent kernel (NTK) theory~\cite{jacot2018neural} views the training of neural network as kernel regression, where the convergence of the network on fitting signals could be derived by analysing the diagonal property of the NTK or the distribution of NTK's eigenvalues. Generally speaking, stronger diagonal property results in better shift-invariance and better convergence, more larger eigenvalues leads to faster convergence for high-frequency components~\cite{tancik2020fourier,bai2023physics}.

Without loss of generality, we focus on a simple case, \textit{i.e.}, the signal to be learned has 1D input and 1D output and FINER has 1 hidden layer with $n$ neurons, such a network could be written as $f(x;\theta)=\sum_{k=1}^{n}c_{k}\sigma(w_{k}x+b_{k})$, where $\sigma(x)=\sin ((|x|+1)x)$ is the activation function. According to the definition, the NTK of FINER could be calculated as\footnote{Please refer the supplemental material for details of derivation.},
\vspace{-0.3cm}
\begin{equation}
\begin{aligned}
        &\mathbf{K}(x_i,x_j)
        =\mathbb{E}_{\theta}\left< \nabla_{\theta}f(x_i;\theta), \nabla_{\theta}f(x_j;\theta) \right>\\
        =&(x_ix_j+1)\mathbb{E}_{\theta}\sum_{k=1}^{n}c_{k}^{2}\underbrace{(2|g_{k}(x_i)|+1)(2|g_{k}(x_j)|+1)}_{\text{Scale term}}\\
        &\underbrace{\cos{((|g_{k}(x_i)|+1)g_{k}(x_i))}\cos{((|g_{k}(x_j)|+1)g_{k}(x_j))}}_{\text{Sign term}}\\
        &\text{where}\:\: g_{k}(x_i)=w_{k}x_i+b_k.
\end{aligned}
\label{eqn:NTK_cal}
\end{equation}
It is observed that, the scale term is approximately proportional to the absolution of bias $b_k$ for all nodes of the kernel, however the change rule of sign term for diagonal elements differs significantly from non-diagonal elements. Specifically, the sign term is always a positive value for diagonal elements while could be either positive or negative for non-diagonal elements.

As a result, the diagonal elements $\mathbf{K}(x_i,x_i)$ are increased with the increase of $|b|$, while the non-diagonal elements $\mathbf{K}(x_i,x_j)$ can be very small, appearing as a diagonal enhanced kernel. According to \cite{jacot2018neural,tancik2020fourier}, NTK with a stronger diagonal property provides better shift-invariance, \textit{i.e.}, the coordinates in the training set are little coupled with each other during the training process, thus the signal could be better learned.

\begin{figure}[t] 
    \begin{center}
        \includegraphics[width=0.95\linewidth]{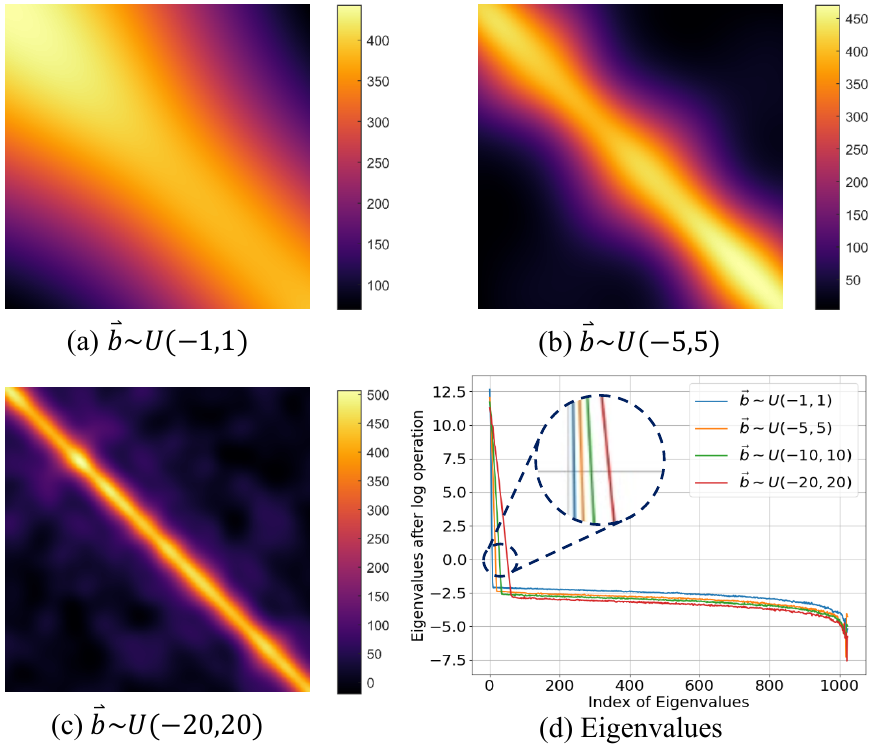}
        \vspace{-0.2cm}
        \caption{Visualizations of NTKs and the corresponding eigenvalues in FINER. (a)-(c) visualize the NTKs when $\vec{b}$ is initialized following $\mathcal{U}(-1,1)$, $\mathcal{U}(-5,5)$ and $\mathcal{U}(-20,20)$, respectively. (d) plots the corresponding eigenvalues. Because the max eigenvalue is much larger than the smallest one, all eigenvalues are processed by a $\log$ function for visualization.}
        \label{fig:NTK_vis}
    \end{center}
    \vspace{-0.7cm}
\end{figure} 

Figs.~\ref{fig:NTK_vis}(a)-(c) visualize the NTKs of FINER for learning a 1D signal with 1024 coordinates. It is observed that the diagonal property of NTK is enhanced with the increase of initialization range of $b$, verifying the analysis mentioned above. Fig.~\ref{fig:NTK_vis}(d) visualizes the changes of eigenvalue distribution, it is observed that the number of eigenvalues which are larger than $10^0$ is significantly increased when larger initialization range is applied to bias. As a result, FINER provides high capacity for learning high-frequency components.

\begin{figure*}[t] 
    \begin{center}
        \includegraphics[width=0.95\linewidth]{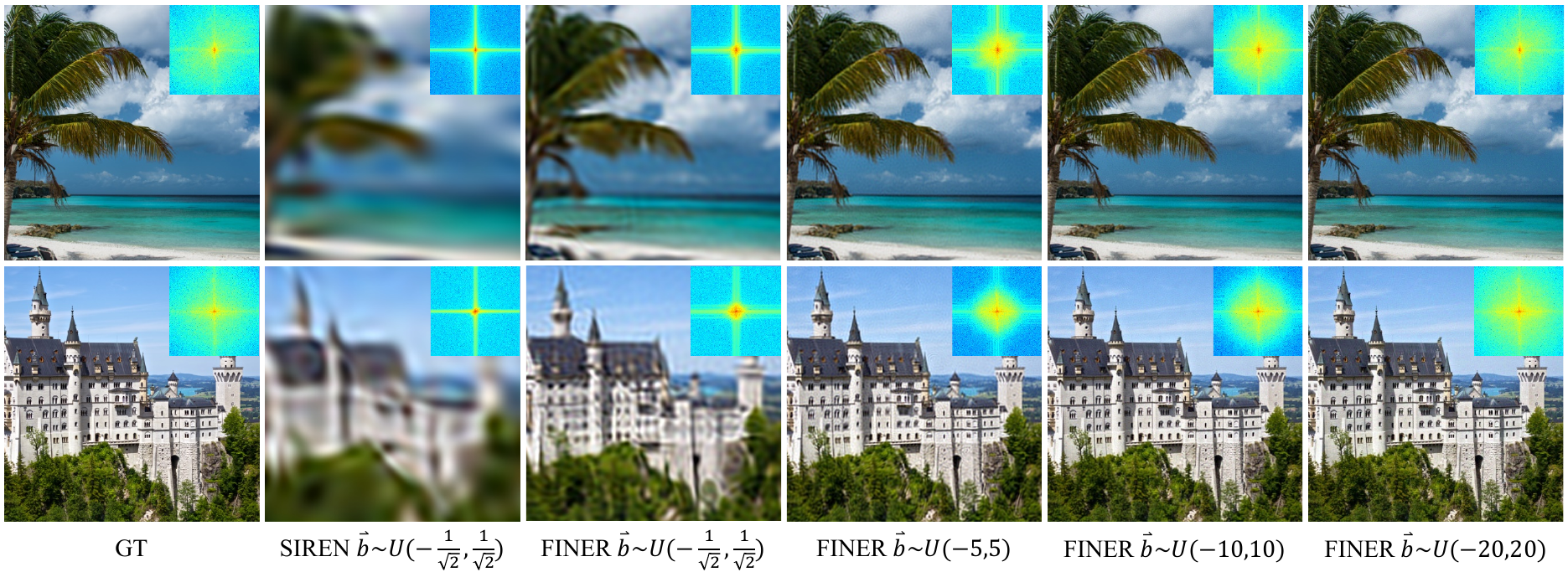}
        \caption{Comparisons of SIREN and FINER with different initializations applied to bias vector $\vec{b}$. For each image, the right-top box visualizes its Fourier spectrum.}
        \label{fig:finer_behavior_vis}
    \end{center}
    \vspace{-0.7cm}
\end{figure*}

\begin{figure}[t] 
    \begin{center}
        \includegraphics[width=\linewidth]{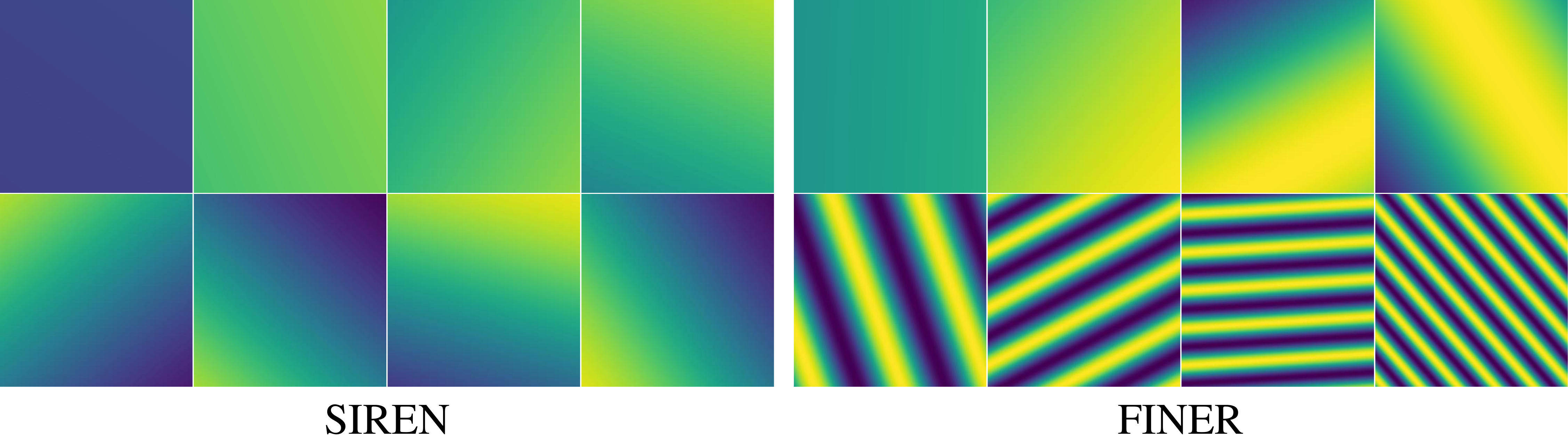}
        \caption{Comparisons of the first layer outputs between SIREN and FINER. For each method, the first row demonstrates 4 neurons with smallest frequencies of the first layer, and the second row refers to 4 neurons with largest frequencies.}
        \label{fig:neurons_cmp}
    \end{center}
    \vspace{-0.7cm}
\end{figure}

\section{EXPERIMENTS}
To verify the behaviors and performance of FINER, three experiments are conducted, including the 2D image fitting, 3D signed distance field representation and the 5D neural radiance fields optimization, where the $\omega_0$ is equal to the one in SIREN and $k$ is set as $\frac{1}{\sqrt{2}}$, $1$ and $\frac{1}{\sqrt{3}}$, respectively.
\subsection{2D Image Fitting}
For the task of 2D image representation, the INR aims at learning a 2D function $f: \mathbb{R}^{2} \rightarrow \mathbb{R}^{3}$ with 2D pixel location input and 3D RGB color output, the loss function is designed as the $L_2$ distance between the network output and the ground truth. To evaluate the performance of INR, the widely used natural dataset~\cite{tancik2020fourier} which contains 16 images with $512\times 512$ resolution is adopted.

\subsubsection{FINER behaviors under different initializations}
\label{sec:img_diff_bias}
To compare the performance of FINER under different initializations better, the pre-set parameter $\omega_0$ is set as $1$ here. According to the analysis in Sec.~\ref{sec:ini_scheme}, the supported frequency set of FINER increases with the initialization range of $\vec{b}$. Different curves in Fig.~\ref{fig:firstimg}(b) reflect this behavior. Fig.~\ref{fig:finer_behavior_vis} visualizes the learned images after $5000$ training epochs. In the 3rd column, because the initialization range of $\vec{b}$ follows the standard method~\cite{sitzmann2020implicit}, FINER has a small supported frequency set, as a result, high-frequency components in the image are lost. With the increase of the initialization range, the supported frequency set of FINER increases and more high-frequency details are provided from the $3$rd column to the last one in Fig.~\ref{fig:finer_behavior_vis}. Additionally, as analysed by Eqn.~\ref{eqn:siren_in_finer}, the supported frequency set of SIREN is a subset of the one in FINER, thus FINER provides more clear details than SIREN when same initializations are applied, \textit{i.e.}, the $2$nd column \textit{vs} the $3$rd column in Fig.~\ref{fig:finer_behavior_vis}. 

According to \cite{yuce2022structured}, the first layer of SIREN or FINER plays the role of frequency encoding. We visualize 8 neurons outputs from total 256 neurons in first layer from SIREN and FINER for comparison, where 4 neurons in the 1st row have the smallest frequencies and the last 4 neurons have the largest frequencies (see Fig.~\ref{fig:neurons_cmp}). It is observed that different neurons in SIREN encode similar frequencies, resulting in a waste of neurons. On the contrary, different neurons in FINER focus on different frequencies, therefore, better representational ability is achieved in FINER.

\definecolor{red}{rgb}{1.0000,0.5686,0.5059}
\definecolor{orange}{rgb}{1.0000,0.7373,0.5059}
\definecolor{yellow}{rgb}{1.0000,0.8431,0.5059}
\begin{table}
\footnotesize
\centering
\caption{Quantitative comparisons on image fitting. We color code each cell as \colorbox{red}{best}, \colorbox{orange}{second best}, and \colorbox{yellow}{third best}.}
\begin{tabular}{lccccc}
\toprule
 {Metrics}
& {PEMLP} & {Gauss} & {SIREN} & {WIRE} & {FINER} \\
\midrule
PSNR $\uparrow$ &29.60 & \cellcolor{yellow}35.39  & \cellcolor{orange}38.52 & 31.31 & \cellcolor{red}40.76\\
SSIM $\uparrow$ &0.8484 & \cellcolor{yellow}0.9455 & \cellcolor{orange}0.9724 & 0.8738 & \cellcolor{red}0.9790\\
LPIPS $\downarrow$ & 1.21e-1 & \cellcolor{yellow}1.91e-2 & \cellcolor{orange}5.52e-3 & 6.45e-2 & \cellcolor{red}2.56e-3\\
\bottomrule
\end{tabular}
\label{tab:res_img}
\vspace{-0.5cm}
\end{table}

\begin{figure*} 
    \begin{center}
        \includegraphics[width=0.95\linewidth]{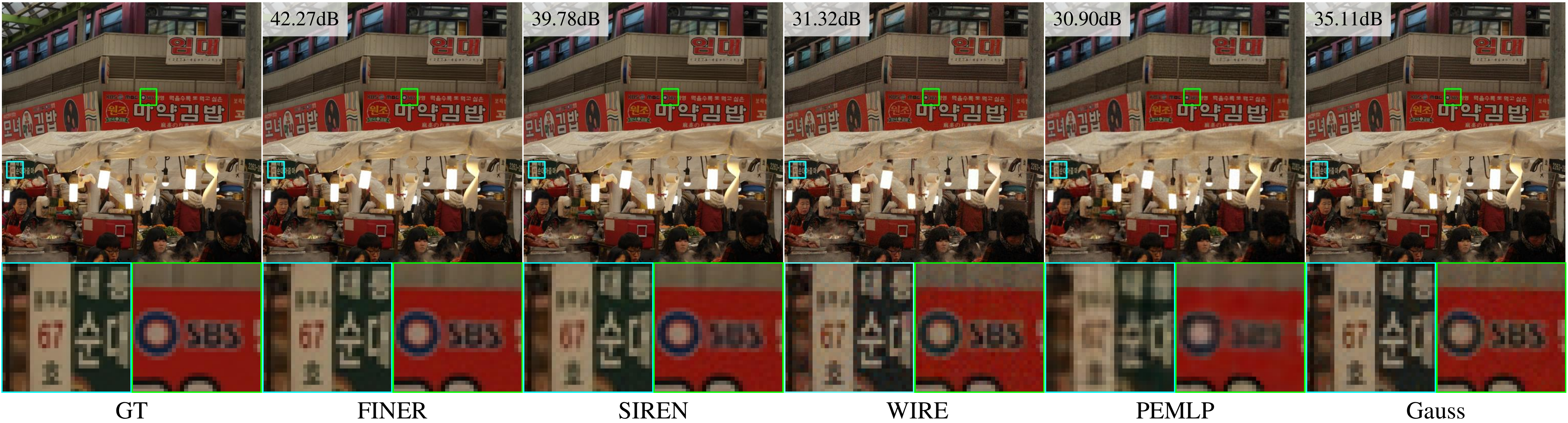}
        \vspace{-0.2cm}
        \caption{Qualitative comparisons between the FINER and baselines on fitting images.}
        \label{fig:res_img}
    \end{center}
    \vspace{-0.7cm}
\end{figure*}
\begin{figure*}[!t] 
    \begin{center}
        \includegraphics[width=0.95\linewidth]{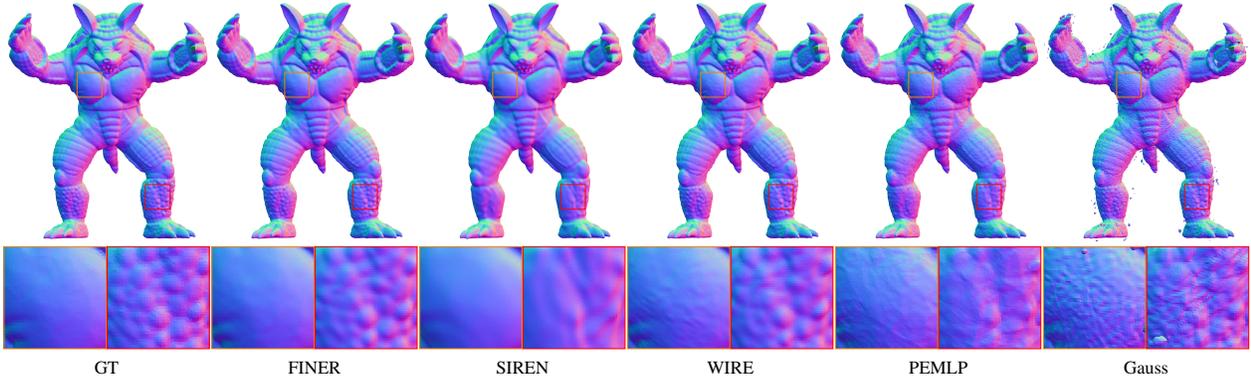}
        \vspace{-0.3cm}
        \caption{Qualitative comparisons on representing the signed distance field of Armadillo.}
        \label{fig:res_sdf}
    \end{center}
    \vspace{-0.7cm}
\end{figure*} 

\subsubsection{Comparisons with the State-of-the-arts}
We compare FINER with four classical INRs, \textit{i.e.}, the Fourier feature embedding (PEMLP)~\cite{mildenhall2020nerf}, INR with periodic activation functions (SIREN)~\cite{sitzmann2020implicit}, Gaussian activation functions (Gauss)~\cite{ramasinghe2022beyond} and wavelet activation functions (WIRE)~\cite{saragadam2023wire}. For a fair comparison, all INRs are set with a same network configuration (3 hidden layers with 256 neurons per layer) and trained with the same Adam optimizer~\cite{kingma2015adam} and $L_2$ loss function between the network output and the ground truth, other parameters are set according to the open-source codes released by authors. Tab.~\ref{tab:res_img} compares FINER with others quantitatively. FINER outperforms other method in all three metrics. Fig.~\ref{fig:res_img} demonstrates the details of different methods. Among all 5 methods, FINER provides more clear results such as the texts `67' and `SBS' in the cyan and green boxes, respectively. On the contrary, these texts are over-smoothed in the results of SIREN and PEMLP. Although WIRE and Gauss could also provide clear texts here, unwelcome serious artefacts also appear in the smooth backgrounds, such as white and red billboards.

\subsection{3D Shape Representation}
Signed distance field (SDF) is one of the most commonly used implicit surface representations in the computer graphics~\cite{jones2006distance}. As the name implies, SDF characterizes the distance from the given 3D point to the closest surface using a continuous function, and the sign of the distance is used to denote whether the point is inside (negative) or outside (positive) the surface. Recently, representing the SDF using INR is drawing more and more attentions~\cite{li2023neuralangelo}. Given a 3D point $\vec{x}$, INR learns a 3D mapping function $f: \mathbb{R}^{3} \rightarrow \mathbb{R}^{1}$ to output the signed distance field values $s$. 
 We apply the FINER to this task and compare to four classical INRs mentioned above. 
 In the experiment, 5 shapes from public dataset~\cite{standord-3D-scanning,muller2022instant} 
 are used for evaluation. For a fair comparison, all methods use a same network configuration, \textit{i.e.}, 3 layers with 256 neurons per layer, additionally, the same coarse-to-fine loss function is used according to \cite{lindell2021bacon}. In the training stage, 10k points are randomly sampled in each iteration and is repeated $200k$ iterations. In the testing stage, a $512^3$ grid is extracted for evaluation and visualization.

\begin{table}
\setlength\tabcolsep{2.5pt}
\scriptsize
\centering
\caption{Quantitative comparisons on representing signed distance field. 
}
\begin{tabular}{clcccccc}
\toprule
 &{Methods}
& {Armadillo} & {Dragon} & {Lucy} & {Thai Statue} 
& {BeardedMan} & {Avg.} \\
\midrule
\multirow{5}*{\rotatebox{90}{Chamfer $\downarrow$}} 
& {PEMLP} & 3.559e-6 & \cellcolor{red}2.081e-6 & \cellcolor{orange}2.224e-6 & 5.284e-6 & \cellcolor{yellow}4.058e-6 & 3.441e-6\\
&{Gauss}  & 1.778e-5 & 7.427e-6 & 5.494e-6 & 1.618e-5 & 1.620e-5 & 1.262e-5 \\
&{SIREN}  & \cellcolor{yellow}3.505e-6 & 2.759e-6 & 2.493e-6  & \cellcolor{yellow}4.481e-6 & \cellcolor{red}3.952e-6  & \cellcolor{yellow}3.438e-6  \\
&{WIRE}  & \cellcolor{red}3.346e-6  & \cellcolor{orange}2.101e-6 & \cellcolor{yellow}2.238e-6 & \cellcolor{orange}3.979e-6  & 4.597e-6  & \cellcolor{orange}3.252e-6 \\
&{FINER}  & \cellcolor{orange}3.348e-6 & \cellcolor{yellow}2.364e-6 & \cellcolor{red}2.119e-6  & \cellcolor{red}3.580e-6  & \cellcolor{orange}4.023e-6 & \cellcolor{red}3.087e-6  \\
\midrule
\multirow{5}*{\rotatebox{90}{IOU $\uparrow$}} 
&{PEMLP} & 9.870e-1 & \cellcolor{red}9.750e-1 & \cellcolor{red}9.760e-1 & \cellcolor{yellow}9.526e-1 & \cellcolor{yellow}9.939e-1 &  \cellcolor{orange}9.769e-1\\          
& {Gauss}  & 9.768e-1 & 9.679e-1 & 9.601e-1 & 9.481e-1 & 9.932e-1 & 9.692e-1\\         
& {SIREN}  & \cellcolor{orange}9.895e-1 & 9.666e-1 & \cellcolor{yellow}9.721e-1 & 9.514e-1 & \cellcolor{red}9.948e-1 &  9.749e-1\\
& {WIRE}   & \cellcolor{yellow}9.893e-1 & \cellcolor{yellow}9.723e-1 & 9.707e-1  & \cellcolor{orange}9.565e-1 & 9.911e-1 &  \cellcolor{yellow}9.760e-1\\  
& {FINER}  & \cellcolor{red}9.899e-1 & \cellcolor{orange}9.725e-1 & \cellcolor{orange}9.756e-1 & \cellcolor{red}9.625e-1 & \cellcolor{orange}9.943e-1 &  \cellcolor{red}9.790e-1\\        
\bottomrule
\end{tabular}
\label{tab:res_sdf}
\vspace{-0.5cm}
\end{table}

\begin{figure*}[!t] 
    \begin{center}
        \includegraphics[width=0.95\linewidth]{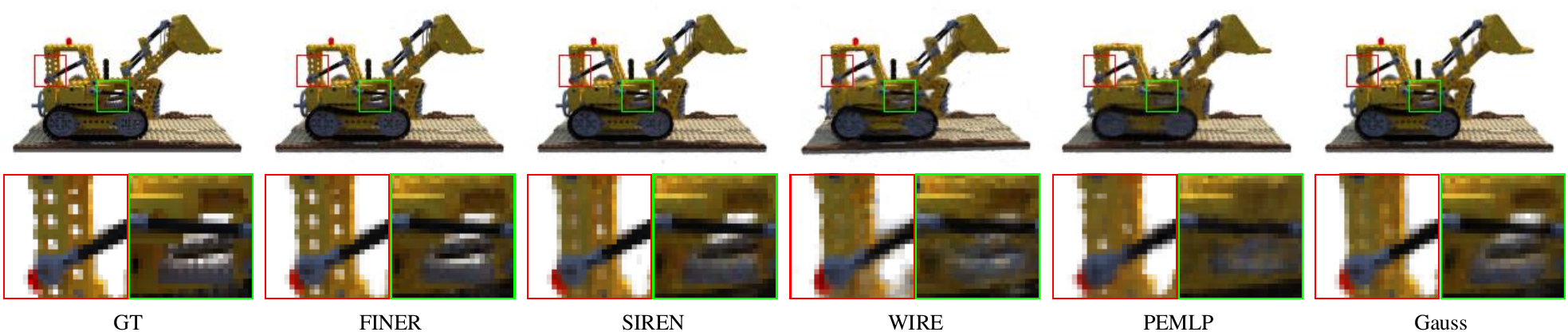}
        \vspace{-0.3cm}
        \caption{Qualitative comparisons between the FINER and baselines on NeRF.}
        \label{fig:res_nerf}
    \end{center}
    \vspace{-0.7cm}
\end{figure*}

 Tab.~\ref{tab:res_sdf} provides quantitative comparisons between the proposed FINER and four baselines. Because FINER provides more freedom for tuning the supported frequency set, best results are achieved among all methods. Fig.~\ref{fig:res_sdf} compares the reconstructed details visually on the Armadillo rendered using Marching Cubes~\cite{lorensen1998marching}. Two representative regions, \textit{i.e.}, the low-frequency smooth pectoral and the high-frequency rough shank, are zoomed-in for comparisons. SIREN represents the smooth pectoral well but loses the details of the shank, WIRE overcomes the limitation of SIREN for representing high-frequency shank at the cost of rough pectoral. For PEMLP, because the pre-defined frequency may not match the frequency distribution in the SDF of Armadillo, both the smooth pectoral and rough shank are not well represented. Guass could not provide stable representation for all shapes (Tab.~\ref{tab:res_sdf}) and there are obvious artefacts in the reconstructed SDF such as the noise outside the shape in Fig.~\ref{fig:res_sdf}. Compared with these baselines, FINER could provide consistent performance for reconstructing both the low- and high-frequency components.

\subsection{Neural Radiance Fields Optimization}
Novel view synthesis, which aims at rendering realistic images at uncaptured poses from a set of images captured at different positions, is one of the key problems in both communities of computer vision and graphics. Recently, representing scenes as neural radiance fields (NeRF)~\cite{mildenhall2020nerf} using INR dominates this task due to the advantages of realism and scalability for embedding different rendering processes~\cite{tewari2022advances}. Given a 3D point 
$\vec{x}$ and a 2D observed direction $\vec{d}$, NeRF focuses on learning a 5D mapping function $f: \mathbb{R}^{5} \rightarrow \mathbb{R}^{4}$ with 5D $(\vec{x}, \vec{d})$ to its 3D color $c$ and 1D opacity $\sigma$. For any pixel $p$ in novel view images, its ray function in 3D space is firstly calculated using the in/extrinsic matrices of camera~\cite{hartley2003multiple}, then $N$ points are sampled along the ray within a predefined depth range, furthermore the direction and position coordinates of these $N$ points are fed into the INR for querying the radiance value $(c,\sigma)$, finally the color $C(p)$ of the pixel is calculated using the differentiable volume rendering technique~\cite{max1995optical,mildenhall2020nerf},
\begin{equation}
\label{eqn:exp_nerf}
\begin{aligned}
C(p)&=\sum_{i=1}^N T_i\left(1-\exp \left(-\sigma_i \delta_i\right)\right) c_i \\
T_i&=\exp \left(-\sum_{j=1}^{i-1} \sigma_j \delta_j\right),
\end{aligned}
\end{equation}
where $\delta_i$ represents the distance between the neighbors in the sampled $N$ points.

We evaluate the FINER for this task and compare to four classical methods mentioned above. To better verify the advantages of FINER for representing high-frequency components, we follow the experimental setting of WIRE that only 25 images are used for training instead of commonly used 100 images, additionally the same $\omega_0$ is set in both FINER and SIREN. Tab.~\ref{tab:res_nerf} and Fig.~\ref{fig:res_nerf} provide quantitative and qualitative comparisons of FINER against different methods on the Blender dataset~\cite{mildenhall2020nerf}. FINER achieves the best performance in almost all 8 scenes. Fig.~\ref{fig:res_nerf} demonstrates the advantage of FINER for representing high-frequency components visually. For example, the holes (red boxes) and the highlights (green boxes) in the frame of the truck are over-smoothed in the reconstructions of all baselines, only FINER could provide clear details here.

\begin{table}
\setlength\tabcolsep{2.5pt}
\scriptsize
\centering
\caption{Quantitative comparisons on novel view synthesis.}
\begin{tabular}{clcccccccc}
\toprule
 &{Methods}
& {Chair} & {Drums} & {Ficus} & {Hotdog} 
& {Lego} & {Materials} & {Mic} & {Ship} \\
\midrule
\multirow{5}*{\rotatebox{90}{PSNR $\uparrow$}} 
& {PEMLP} & 31.32 & 20.18 & 24.49 & 30.59
                & 25.90 & 25.16 & 26.38 & 21.46 \\
&{Gauss}  & \cellcolor{yellow}32.68 & \cellcolor{yellow}23.16 & \cellcolor{yellow}26.10 & \cellcolor{yellow}32.17
                & \cellcolor{yellow}28.29 & \cellcolor{yellow}26.19 & \cellcolor{orange}33.59 & \cellcolor{orange}22.28 \\
&{SIREN}  & \cellcolor{orange}33.31 & \cellcolor{orange}24.89 & \cellcolor{orange}27.26 & \cellcolor{orange}32.85
                & \cellcolor{orange}29.60 & \cellcolor{red}27.13 & \cellcolor{yellow}33.28 & \cellcolor{yellow}22.25 \\
&WIRE       & 29.31 & 22.22 & 25.91 & 30.11 & 25.76 & 25.05 & 32.35 & 21.15 \\                

&{FINER}   & \cellcolor{red}33.90 & \cellcolor{red}24.90 & \cellcolor{red}28.70 & \cellcolor{red}33.05
                & \cellcolor{red}30.04 & \cellcolor{orange}27.05 & \cellcolor{red}33.96 & \cellcolor{red}22.47 \\
\midrule
\multirow{5}*{\rotatebox{90}{SSIM $\uparrow$}} 
&{PEMLP} & 0.960 & 0.814 & 0.914 & 0.945
                & 0.904 & 0.909 & 0.960 & 0.754 \\
                
& {Gauss}  & \cellcolor{yellow}0.967 & \cellcolor{yellow}0.883 & \cellcolor{yellow}0.933 & \cellcolor{yellow}0.956
                & \cellcolor{yellow}0.932 & \cellcolor{yellow}0.916 & \cellcolor{orange}0.980 & \cellcolor{yellow}0.782 \\
                
& {SIREN}  & \cellcolor{orange}0.971 & \cellcolor{red}0.912 & \cellcolor{orange}0.947 & \cellcolor{red}0.960
                & \cellcolor{orange}0.948 & \cellcolor{red}0.932 & \cellcolor{yellow}0.979 & \cellcolor{orange}0.788 \\
& WIRE      & 0.938 & 0.858 & 0.931 & 0.938 & 0.886 & 0.898 & 0.978 & 0.734\\                
& {FINER}   & \cellcolor{red}0.973 & \cellcolor{orange}0.911 & \cellcolor{red}0.958 & \cellcolor{orange}0.959
                & \cellcolor{red}0.951 & \cellcolor{orange}0.928 & \cellcolor{red}0.981 & \cellcolor{red}0.792 \\                
\midrule
\multirow{5}*{\rotatebox{90}{LPIPS $\downarrow$}} 
& {PEMLP} &0.026 &0.185 &0.056 & 0.037
                &0.070 &0.050 &0.057 & 0.190 \\
                
&{Gauss}  &\cellcolor{yellow}0.019 &\cellcolor{yellow}0.082 &\cellcolor{yellow}0.055 & \cellcolor{orange}0.032
                &\cellcolor{yellow}0.040 &\cellcolor{yellow}0.037 &\cellcolor{orange}0.013 & \cellcolor{yellow}0.136 \\
                
&{SIREN}  &\cellcolor{orange}0.017 &\cellcolor{red}0.051 &\cellcolor{orange}0.033 & \cellcolor{orange}0.032
                &\cellcolor{orange}0.030 &\cellcolor{red}0.028 &\cellcolor{yellow}0.014 & \cellcolor{orange}0.116 \\
&WIRE       & 0.035 & 0.106 & 0.045 & 0.048 & 0.075 & 0.058 & 0.014 & 0.172\\                
&{FINER}   &\cellcolor{red}0.015 &\cellcolor{orange}0.052 &\cellcolor{red}0.024 & \cellcolor{red}0.033
                &\cellcolor{red}0.025 &\cellcolor{orange}0.032 &\cellcolor{red}0.010 & \cellcolor{red}0.108 \\
\bottomrule
\end{tabular}
\label{tab:res_nerf}
\vspace{-0.5cm}
\end{table}

\section{Conclusion}
We have proposed and verified the FINER which uses variable-periodic functions for activating the INR. We have pointed out that current INRs suffer from a limited supported frequency set due to the under-utilization of the definition domain of activation functions. The proposed FINER overcomes this problem by introducing variable-periodic activation function and initializing the bias vector to different ranges, where different sub-functions with different frequencies along the definition domain will be selected for activation. As a result, the supported frequency set of FINER could be flexibly tuned, and the capacity of INR could be significantly enlarged. We have demonstrated the advantages of FINER over other INRs in image fitting, 3D shape representation and neural rendering.


{
    \small
    \bibliographystyle{ieeenat_fullname}
    \bibliography{main}
}


\end{document}